\documentclass[10pt,letterpaper]{article}

\usepackage{cogsci}
\usepackage{booktabs}       % professional-quality tables
\usepackage[pdftex]{graphicx}
\cogscifinalcopy % Uncomment this line for the final submission 

\usepackage[
    backend=biber,
    style=apa,
    natbib=true,
    doi=false,
    isbn=false,
    url=false,
]{biblatex}
\usepackage{pslatex}
\usepackage{float} % Roger Levy added this and changed figure/table
\title{Predicting Word Learning in Children from \\ the Performance of Computer Vision Systems}

\author{%
   \hspace{-5mm} Sunayana Rane\\
  \hspace{-5mm}Department of Computer Science\\
  \hspace{-5mm}Princeton University\\
  \hspace{-5mm}\texttt{srane@princeton.edu} \\
   \And
   \hspace{-20mm}\textbf{Mira L. Nencheva} \\
   \hspace{-20mm}Department of Psychology \\
  \hspace{-20mm}Princeton University\\
   \hspace{-20mm}\texttt{nencheva@princeton.edu} \\
    \And
   \hspace{-10mm}\textbf{Zeyu Wang} \\
   \hspace{-10mm}Department of Electrical and Computer Engineering \\
  \hspace{-10mm}Princeton University\\
   \hspace{-10mm}\texttt{zeyuwang@princeton.edu} \\
    \AND
  \hspace{-5mm}\textbf{Casey Lew-Williams} \\
   \hspace{-5mm}Department of Psychology \\
  \hspace{-5mm}Princeton University\\
   \hspace{-5mm}\texttt{caseylw@princeton.edu} \\
    \And
  \hspace{-20mm}\textbf{Olga Russakovsky} \\
   \hspace{-20mm}Department of Computer Science \\
  \hspace{-20mm}Princeton University\\
   \hspace{-20mm}\texttt{olgarus@princeton.edu} \\
    \And
  \hspace{-10mm}\textbf{Thomas L. Griffiths} \\
  \hspace{-10mm}Departments of Psychology
and Computer Science \\
  \hspace{-10mm}Princeton University\\
   \hspace{-10mm}\texttt{tomg@princeton.edu} \\ 
}

\begin{document}

\maketitle

\begin{abstract}
For human children as well as machine learning systems, a key challenge in learning a word is linking the word to the visual phenomena it describes. We explore this aspect of word learning by using the performance of computer vision systems as a proxy for the difficulty of learning a word from visual cues. We show that the age at which children acquire different categories of words is correlated with the performance of visual classification and captioning systems, over and above the expected effects of word frequency. The performance of the computer vision systems is correlated with human judgments of the concreteness of words, which are in turn a predictor of children's word learning, suggesting that these models are capturing the relationship between words and visual phenomena. 
\end{abstract}

\section{Introduction}

Both humans and machines face the problem of establishing the relationship between visual and linguistic information. In humans, this process is known as word learning, and has been extensively studied by developmental scientists. In machines, linking visual features with words is a key part of several tasks studied by computer vision researchers, including object classification and image captioning.  In this paper, we explore the extent to which the solutions to these problems found by humans and machines are related by predicting the time course of word learning in human children from the performance of computer vision systems.

Developmental scientists have long been interested in understanding how infants and young children learn new words \citep{brown1973first, golinkoff2000becoming, wojcik2022map, bloom2002children, quine1960word}, often framing the problem as one of establishing reference between words and their corresponding objects, events, or properties \citep{markman1990constraints, schwab2016repetition}. While the trajectory of word learning varies across children, there is at least some consistency in the rates at which different kinds of words are learned \citep{frank2021variability}. For example, children learning English (as well as many other languages) tend to learn words describing body parts (such as ``eye'' or ``nose'') earlier than they learn connecting words (such as ``and'' or ``because''). Developmental scientists have looked for predictors of this pattern. For example, words that are more frequent in child-directed speech tended to be learned earlier \citep{swingley2018quantitative}. However, the investigation of these predictors has been limited to quantities that can be measured from linguistic input (such as word frequency) or by adults making an intuitive judgment about the properties of words (such as a word's ``concreteness'' or ``abstractness''). 

Previous work has not made use of predictors that directly measure the correspondence between a word and the visual phenomena it describes. Visual aspects of reference pose a challenge for the child learner because scenes vary in complexity \citep{quine1960word} and because the referents of words can be highly variable (e.g., ``dog'' can refer to both chihuahuas and Bernese mountain dogs). Relatedly, some words and word categories refer to concrete objects (e.g., ``cup''), but others do not (e.g., ``more'' or ``fine''), a dimension known to shape age of acquisition (AoA) \citep{bergelson2013acquisition, swingley2018quantitative}. Prior experimental work has begun to understand how visual context supports word learning. For example, young children can track word/object co-occurrence statistics over time to disambiguate the meanings of novel labels in complex visual scenes \citep{chen2018learning, yurovsky2013statistical}, and infants who see more variable views of objects show more rapid vocabulary growth later on \citep{slone2019self}. Although the ease with which a word can be mapped to a concrete visual referent affects children's noun learning, developmental scientists have not formalized how the infant mind may process and create representations of the statistics of visual scenes and labels. 
 
In this paper, we investigate whether we can capture the visual difficulty of learning words by examining the performance of classification and image captioning systems. Since these systems need to solve similar problems to children, they may face the same difficulties. We look at how well object classification and image captioning systems perform for different categories of words (such as animals vs. furniture), and use the resulting performance measures to predict children's word learning. Our results show -- across different tasks and architectures -- that the difficulty with which machines learn words in different categories is a good predictor of the difficulty with which children learn words in those categories, and that including this measure improves prediction of children's word learning over just using word frequency. We also show that the performance of the computer vision systems is correlated with human judgments of the ``concreteness'' of a word, which is known to predict AoA. Computer vision models thus provide an automated measure of this subjective quantity.
  
While human children and deep neural networks for object classification and image captioning are presumably quite different kinds of systems, discovering parallels in their performance suggests that some aspects of the difficulty with which different kinds of words are learned is a consequence of the nature of the problem itself. Just as the statistics of the linguistic input to children play a key role in understanding language acquisition \citep{montag2015words,laing2020babble,bergelson2017nature}, the statistics of correspondences between that linguistic input and the world that it describes are significant. Our results demonstrate how improvements in computer vision systems offer new opportunities for the scientific study of child development.

%An important part of word learning is relating a word to the visual phenomena it describes. Deep features from computer vision systems \citep{lecun2015deep} are known to predict human perceptual similarity judgements with surprising accuracy \citep{human-cnn}. However, we don't yet know whether deep features from computer vision and language models also capture the visual difficulty of learning a word. We use visual classification and captioning systems as a proxy for the difficulty of learning a word from visual cues, and show that the performance of these models predict the age at which children acquire different categories of words. 

%\subsection{Background}
%\subsubsection{Image classification and image captioning}
%[Olga/Sunayana]
%\subsubsection{Predicting children's word learning}

% A better understanding of how effectively vision and language models compare to child word learning behavior has a twofold benefit: first, it helps us model child word learning and understanding of increasingly complex visual concepts. Second, it enables us to understand models' developmental parallels to their human counterparts, and create models with more human-like behaviors. This work is a step towards that goal, with a behavioral analysis showing similarities of visual word learning in models and children.   

\section{Datasets and models}
To investigate model word learning in a way that is relevant to child word learning, we need two kinds of data: 
\begin{enumerate}
    \item Child word learning data, including which words (or word categories) are learned at various developmental stages.
    \item Standardized image and natural language data,  which can be used to train vision and language models to produce language comparable to child word production.
\end{enumerate}
We address both these needs by working with multiple sources of data: WordBank for child word learning, and COCO for training our models. 
\subsection{Child language acquisition data: WordBank}
\begin{figure}[ht]
\begin{center}
\includegraphics[width=0.50\textwidth,height=0.28\textwidth]{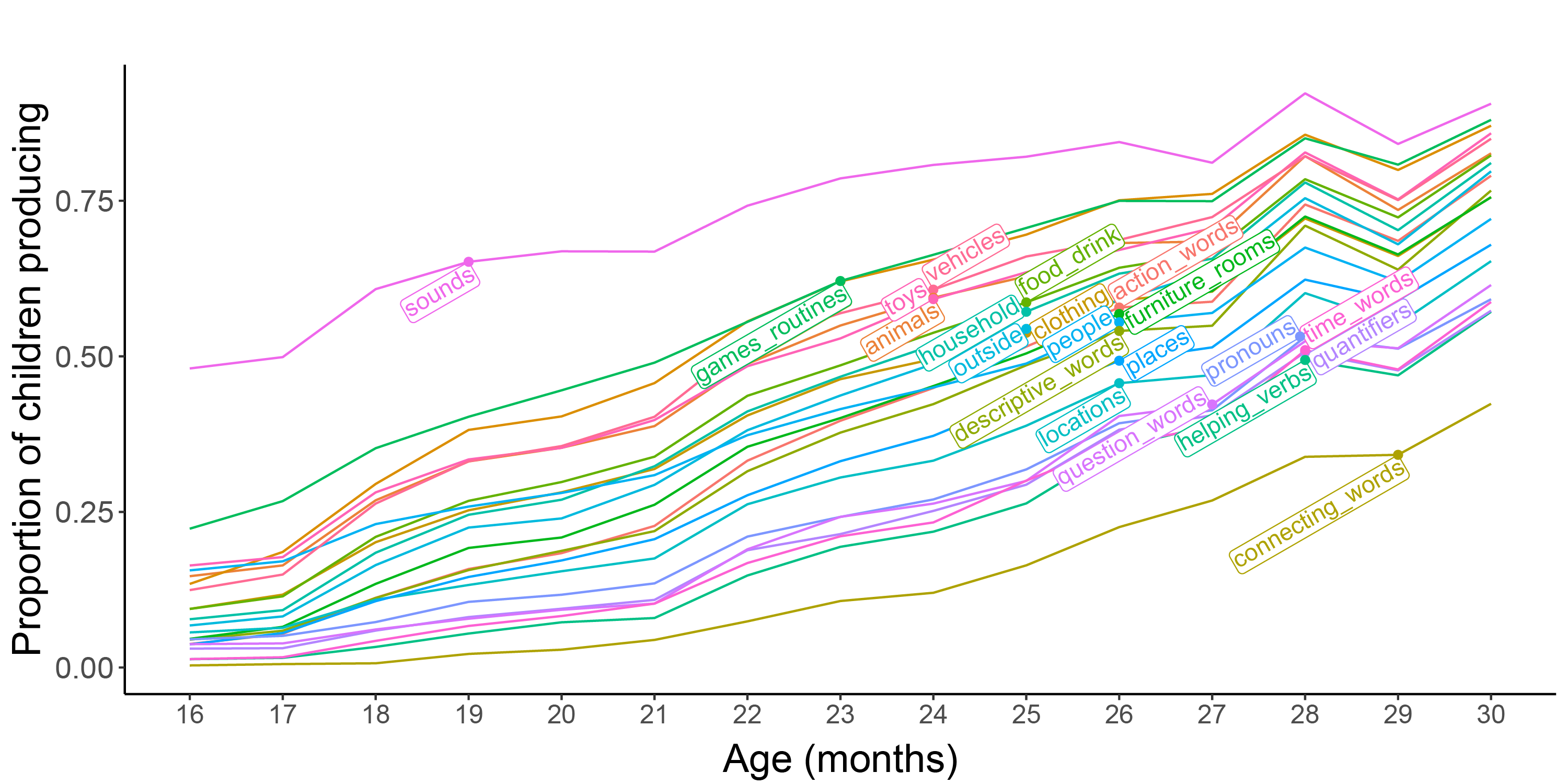}
\end{center}
\caption{\label{fig:wordbank} WordBank data tracking  production of different categories across different ages. Each line represents the average proportion of children producing all words in a given category on the MCDI. The age at which half of children produce a given word (age of acquisition; AoA) is marked with a circle on each line. We aim to predict the order of acquisition of words in these categories using the performance of computer vision models for the corresponding words.}
\end{figure}

We use data sourced from the WordBank child language database \citep{wordbank} to extract words commonly produced by toddlers between the ages of 16 and 30 months. Figure~\ref{fig:wordbank} gives an example of the type of data collected and tracked by WordBank. In particular, the data we use corresponds to which words are easily (and not easily) produced by toddlers of various ages. WordBank contains production percentiles for approximately 1200 words, which we use for our analysis of model word production. %Below, we detail why we also use WordBank word categories as the level of granularity for this analysis.

% Cogsci revision (removed): Figure~\ref{fig:wordbank} gives an example of the type of data collected and tracked by WordBank. In particular, the data we use corresponds to which words are easily (and not easily) produced by toddlers of various ages.

In order to compare word learning at scale, we decided to investigate patterns in word categories instead of individual words. For an effective comparison to child word learning, we use word categories for which there exists parallel child data. Fortunately, the WordBank database contains such data. We extracted approximately 1200 frequently-produced words for toddlers, as listed in the WordBank database, and mapped them to the corresponding category. WordBank categories include people, toys, animals, etc. We remove sounds/sound effects (such as “cockadoodledoo”) from these categories, because our models are restricted to vision and language.

%are rarely trained on onomatopoeia of textual forms of sound effects of any kind (a necessary tradeoff when limiting to the visual and linguistic modalities in order to constrain the problem and provide a meaningful analysis). 

\subsection{Computer vision tasks: COCO}

On the computer vision side we use the canonical COCO~\citep{COCO} image captioning dataset (the Karpathy split, to be precise~\citep{karpathy}) for training and evaluating the models. The dataset contains 113,278 training and 5,000 validation images, each associated with 5 captions provided by human annotators. We use two computer vision tasks in our experiments: image captioning, where a model is trained to produce a natural language caption, as well as the simpler image classification task, where a model is trained to predict which visual categories are present in the image, without tying these categories to a natural language description.
%The first is the more complex image captioning task, where a model is trained to emulate the natural language caption and then is evaluated on the similarity of that caption with the manual captions. The second is the simpler image classification task, where a model is trained to predict which visual categories are present in the image, without tying these categories together into a natural language description. %The image classification setup allows for a more direct evaluation of the visual difficulty of each concept, albeit at the cost of being further from emulating human communication. 
For image classification, we create (imperfect) binary labels comprising of 855 individual words which are in both COCO and WordBank. The binary label is determined by whether the word is mentioned in one of the captions associated with the image.    

%use the manual captions to create a binary multi-label target using 855 visual categories present in WordBank. 

%To adhere to the standard training protocols for captioning models, we use the canonical image captioning dataset,  COCO captions \citep{coco} for training these models. For the classification models, we use each image's COCO captions to derive multilabel output as follows: for each word in the captions corresponding to an image, we create a binary label. We then restrict this set of labels to only the words which are also present with corresponding child data in WordBank, giving us ~855 categories. We describe this process in detail in the Training and Hyperparameters section for Classification Models. We use the Karpathy train and validation splits over COCO for training our captioning models and report correlation numbers for the validation split \citep{karpathy}. 

%\section{Computer vision models}

%\olga{I suggest merging this above -- so having 2.1 (which includes current 2.1 and 2.2 for WordBank) and then 2.2 (which includes current 2.3 and 3) for computer vision}

We run experiments with canonical computer vision models. Our goal is to verify that our findings hold across a range of standard setups. For image classification we use two different CNN architectures: VGGNet~\citep{vggnet} and ResNet50~\citep{resnet}, both with and without pretraining on ImageNet~\citep{imagenet-challenge}. For image captioning, we explore two more complex vision backbones: a ResNet101 CNN~\citep{resnet} or bottom-up features from Faster R-CNN~\citep{faster,bottomup}. We combine these models with one of two language models: the classic LSTM~\citep{bottomup} or the more recent Transformer~\citep{transformer}. 

We use open-source implementations of all models along with the recommended hyperparameters~\citep{captioning-repo1,captioning-repo2,keras}. Each model is trained on a single GPU. For image classification, we train with an adaptive learning rate using the Adam \citep{adam} optimizer and dropout until the loss converged. VGGNet trained from scratch proved difficult to train to convergence, despite performing grid search over initial learning rate, dropout, and batch size, so only pretrained results are reported. In the captioning models, for the LSTM layers, we use an input encoding size of 1000, a hidden size of 512, a batch size of 10, and an adaptive learning rate. For the Transformer layers, we use 8 attention heads, 6 encoder and decoder layers, 512 hidden unit size, and a batch size of 10. We keep as much consistent as possible between the implementations, so that we can have a meaningful comparison.

\section{Metrics}

To quantitatively measure the correspondence between word learning in human children and computer vision systems, we adopted standard metrics used in the relevant fields: the median age at which children produce a word, and the word-level performance measures of AUC for classifiers and SPICE for captioning systems.

\subsection{Metric for children: AoA (Age of Acquisition)}
The Age of Acquisition (AoA) of a word is defined as the age at which at least 50\% of children produce the word. WordBank includes the vocabularies of 5520 toddlers learning North American English assessed using parent report on the MacArthur Bates Communicative Development Inventory (MCDI)  \citep{fenson2007macarthur}. In the WordBank database, AoA can be calculated over the parent-reported scores for word learning for each child in the database. AoA has previously been shown to correspond with the difficulty of learning to read a word \citep{aoa}. We use this measure of AoA as a proxy for the difficulty of learning a word for a child. AoA was calculated for each word and then averaged within each of the WordBank categories: body parts, animals, vehicles, toys, household, outside, food/drink, furniture/rooms, clothing, locations, descriptive words, places, people, action words, pronouns, question words, quantifiers, helping verbs, time words, and connecting words. %We then compare AoA to corresponding measures that reflect a model's difficulty in learning the same word category, such as AUC for classification models and SPICE for captioning models.  

\subsection{Metrics for machines: AUC and SPICE}

The metrics we use for our models are designed to measure a model's performance at the level of individual words. For classification models, we use AUC (the area under the receiver operating characteristic (ROC) curve \citep{stats-textbook}) as a classification metric which is robust to class imbalance. Our multi-label, multi-class classification task was binary for each label, so a per-word AUC calculated over each label was an appropriate metric. 

For captioning models, we use the Semantic-Propositional Image Caption Evaluation score (SPICE) \citep{spice}. SPICE is an automatic evaluation metric which uses scene graphs corresponding to the actual image to gauge semantic and propositional correctness, instead of just the textual n-gram comparison of previous metrics. 
%TODO - Put back in camera-ready: Scene graphs contain objects, attributes of those objects, and relationships between those objects as represented in an image \citep{scene-graphs}. Recent studies have shown that many standard automated evaluation metrics of captioning models fall short of human judgements \citep{captioning-eval}, including BLEU \citep{bleu} and CIDEr \citep{cider}. However, although all these automated metrics have drawbacks, SPICE has repeatedly been shown to be robust to comparison with human judgements \citep{captioning-eval}. 
From a caption like “woman sitting on a brown chair in a restaurant”, SPICE produces a tuple-based scene graph containing tuples like “woman-sitting” and “sitting-on-chair.” SPICE then calculates whether each produced tuple matches one of the tuples that appear in the ground truth manual captions. To get a score for each individual WordBank word, we then average the tuple-based scores of all the tuples where the word appears. The intuition for using this metric is that it is impossible to gauge whether a word like “sitting” is used correctly without looking at the other words around it. %Derived from an F1-style metric, SPICE is also robust to class imbalance and is similar to AUC, while allowing for more complex analysis over word tuples.

For both AUC and SPICE, after calculating at a word-level (or a tuple-level, for SPICE) we then aggregate over WordBank categories for ease of comparison to AoA for those same categories. 

\section{Results}

As an initial analysis, we examined the raw correlation between AoA and the machine metrics. We then conducted a series of multiple regression analyses to determine whether computer vision systems can improve prediction of the timecourse of child word learning over existing measures used in the child language acquisition literature.

\subsection{Correlations}
\begin{figure*}[ht]
\begin{center}
\includegraphics[width=\textwidth] {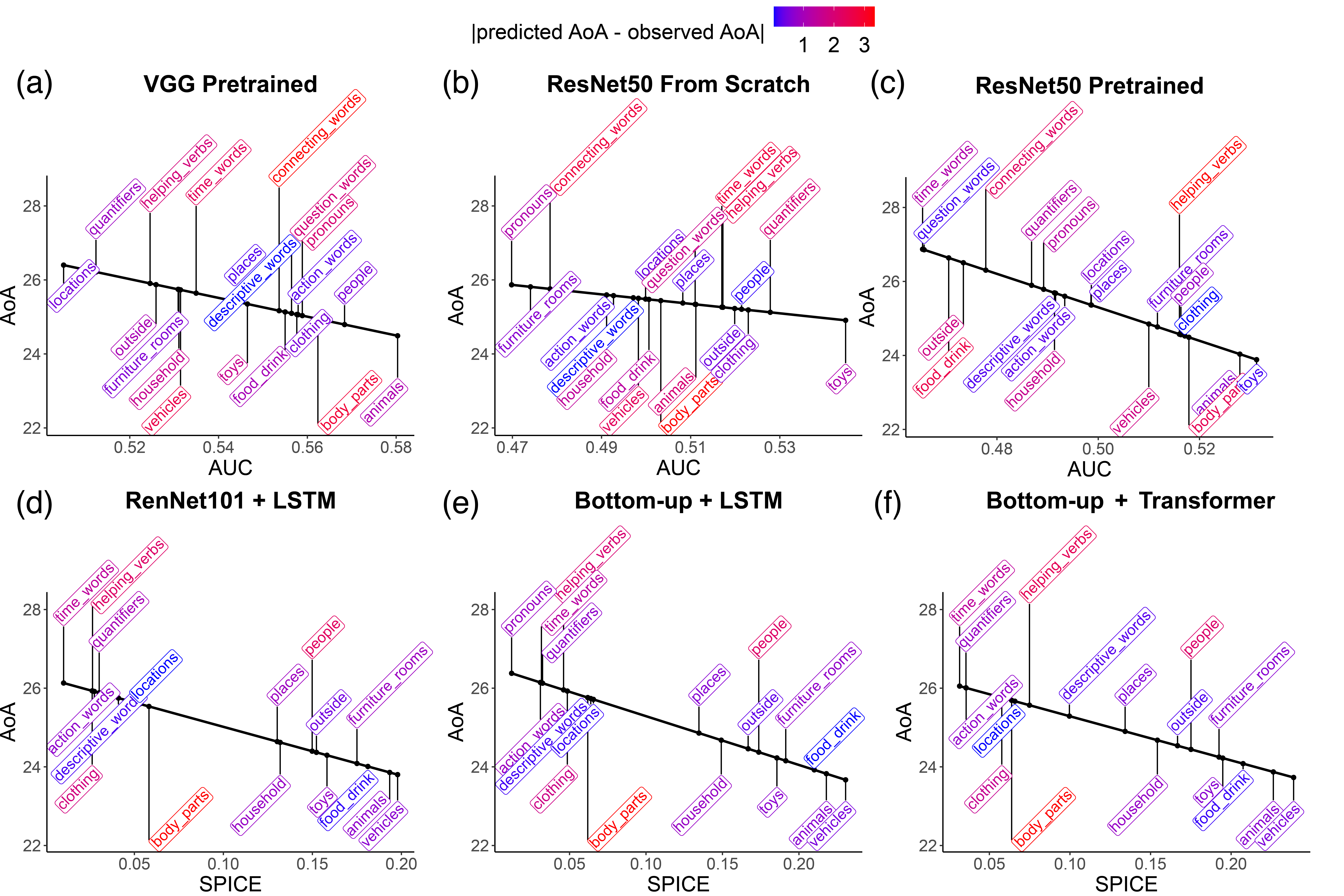} %{regression_plot_v5.png}
\end{center}
\caption{\label{fig:regressions} Regression of successful models' performance vs age of acquisition in months, per category. Each category label is placed so that its bottom left corner indicates the corresponding age of acquisition (AoA) and AUC (classification) or SPICE (captioning) values. The black regression line shows the AoA predicted by each model in the regression, and the vertical residual lines and category labels show the observed average AoA for the category. }
\end{figure*}

To compare the word category-level AoA to AUC/SPICE of the models, we report two types of correlation: the Pearson correlation coefficient, which assumes a linear relationship, and the Spearman rank-order correlation coefficient, which only assumes monotonicity \citep{stats-textbook}. The results are shown in Table \ref{clf-results} for classification and Table \ref{captioning-results} for captioning, with corresponding scatterplots in Figure \ref{fig:regressions}.

\begin{table}
%\resizebox{\textwidth}

\tiny
  \caption{Classification model AUC correlation with AoA}
  \vspace{2mm}
  \label{clf-results}
  \centering
  \begin{tabular}{llllll}
    \toprule
    \multicolumn{2}{c}{Classification model setup}                   \\
    \cmidrule(r){1-2}
 Model & Training &  Pearson & $p$ & Spearman & $p$\\
 \midrule
  VGG  &  Pretrained & -0.280 & 0.232 & -0.311 & 0.182\\
  ResNet50  & From scratch & -0.138 & 0.562 & -0.081 & 0.734\\
  ResNet50  & Pretrained & -0.531 & 0.016 & -0.544 & 0.013\\
    \bottomrule
  \end{tabular}
\end{table}

\begin{table}
\tiny
  \caption{Captioning model SPICE correlation with AoA}
  \label{captioning-results}
  \centering
  \vspace{2mm}
  \begin{tabular}{llllll}
    \toprule
    \multicolumn{2}{c}{Captioning architecture}                   \\
    \cmidrule(r){1-2}
 Pretrained Visual Features & Language Layers &  Pearson  & $p$ & Spearman & $p$ \\
  \midrule
  ResNet 101  &  LSTM & -0.515 & 0.034 & -0.640 & 0.006\\
  Bottom-up (Faster R-CNN)  & LSTM & -0.617 & 0.004 & -0.708 & 0.000 \\
    Bottom-up (Faster R-CNN)  & Transformer & -0.565 & 0.012 & -0.624 & 0.004\\
    \bottomrule
  \end{tabular}
\end{table}

Four of our models showed statistically significant correlations with AoA: classification ResNet50 with pretrained features, and all three of the captioning models. In all four of these models, as performance (AUC or SPICE) increased, AoA decreased: categories of words that were easier for the models were acquired earlier by children. The remaining two classification models (VGG with pretrained features and ResNet50 trained from scratch) also showed correlations consistent with this relationship, but those correlations were not statistically significant. The correlation for ResNet50 trained from scratch was particularly weak, suggesting that pretrained features may be important. The correlations for the captioning systems were all of similar magnitude, suggesting that the specific architecture (including the choice of an LSTM or Transformer) may be less relevant than the captioning task itself.

\subsection{Comparison with other predictors}

As noted above, developmental psychologists have explored variables that predict children's word learning. One such variable is the frequency with which words appear in child-directed speech. We evaluated the correlation between word frequency (extracted from the TalkBank database \citep{talkbank}) and AoA, finding a Pearson correlation of $r=$ -0.377 ($p = 0.092$) and a statistically significant Spearman correlation of $\rho=$ -0.494 ($p = 0.022$). %Cogsci revision: Notably, this correlation is of similar magnitude to the correlations produced by our computer vision metrics. See Appendix for a scatterplot.

To determine whether our computer vision metrics (AUC and SPICE) are predictive of AoA even after accounting for word frequency, we conducted a multiple regression analysis where the independent variables are word frequency and AUC/SPICE, and the dependent variable is AoA. The coefficients of the multiple regression analysis show that AUC/SPICE across different successful models do indeed predict AoA over and above frequency in child-directed data. The results are shown in Table \ref{multiple-regression}. The multiple regression analysis showed that all four predictors that originally resulted in a statistically significant correlation with AoA remained statistically significant when word frequency was taken into account. Notably, word frequency was no longer a statistically significant predictor in the resulting models.
%Our results show a strong negative correlation between AoA and SPICE for all captioning models, and between AoA and AUC for the pretrained  ResNet50 classification model. A negative correlation between AoA and SPICE means that words with a low AoA (acquired earlier, and therefore easier to learn) also have a high SPICE (better model performance). The same holds for AUC: high AUC corresponds to low AoA, meaning the model performance is higher for word categories which are easier for children to learn. 
% Notes from Mira: We found a positive correlation between human concreteness ratings (Köper & im Walde, 2017) and AUC and SPICE

%\begin{table}
%  \caption{Regression coefficient for word frequency in child-directed speech vs AoA, without accounting for any visual phenomena}
%  \label{frequency-regression}
%  \centering
%  \begin{tabular}{lll}
%
%    \toprule
%    \cmidrule{1-2}
% TalkBank frequency & $p$ \\
% \midrule
%   -2.444 & 0.043 \\
%    \bottomrule
%  \end{tabular}
%\end{table}

\begin{table}
\tiny
  \caption{Multiple regression with TalkBank word frequency as a predictor in addition to AUC or SPICE}
  \label{multiple-regression}
  \centering
  \vspace{2mm}
  \begin{tabular}{llllllll}
    \toprule
    \multicolumn{6}{c}{Regression Coefficients (predicting AoA)}                   \\
    \cmidrule(r){1-6}
 Model & TalkBank  & $p$ &  AUC / %(classification) 
 & $p$ & $R^2$\\ &frequency & &SPICE \\ %(captioning)\\
 \midrule
  VGG Pretrained & -1.891 & 0.128 & -1.580 & 0.306 & 0.199\\
  ResNet50 From Scratch & -2.160  & 0.087 & -1.243 & 0.431 & 0.178\\
  ResNet50 Pretrained & -1.286 & 0.269 & -2.559 & 0.044 & 0.333\\
  ResNet101 + LSTM & -1.297 & 0.267 & -2.148 & 0.050 & 0.329\\
  Bottom-up + LSTM & -1.225 & 0.229 & -2.769 & 0.007 & 0.433\\
  Bottom-up + Transformer & -1.552 & 0.153 & -2.355 & 0.027 & 0.404\\
    \bottomrule
  \end{tabular}
\end{table}

Another variable that has been shown to be a good predictor of AoA is the ``concreteness'' of words \citep{swingley2018quantitative, bergelson2013acquisition}.  Unlike word frequency, concreteness is not a property that can be measured directly from the linguistic input to children. Rather, it is typically measured by asking human raters to rate on a scale how  ``concrete'' or ``abstract'' they consider a word to be, typically after providing a definition for concrete (e.g. can be experienced with the five senses) and abstract words (e.g. cannot be experienced through the five senses) \citep{brysbaert2014concreteness}. Some work has expanded these rating lists by using supervised models trained directly to predict concreteness \citep{koper2017improving}.  We ran a second  multiple regression using a standard measure of concreteness (taken from \cite{koper2017improving}) as a predictor. The results are shown in Table \ref{multiple-regression2}. In this model, concreteness was the only statistically significant predictor, with neither word frequency nor our measures being statistically significant. The same result is observed in a multiple regression incorporating only  concreteness and AUC/SPICE as predictors \ref{regression-concreteness}.

\begin{table}
\tiny
  \caption{Multiple regression with concreteness judgments as a predictor in addition to AUC or SPICE}
  \label{regression-concreteness}
  \centering
  \vspace{2mm}
  \begin{tabular}{lllllll}
    \toprule
    \multicolumn{6}{c}{Regression Coefficients (predicting AoA)}                   \\
    \cmidrule(r){1-6}
 Model & AUC & $p$ &  Concreteness  %(classification) 
 & $p$ & $R^2$\\  &SPICE & & \\ %(captioning)\\
 \midrule
  VGG Pretrained & -0.128 & 0.884 & -3.925 & 0.000 & 0.754\\
  ResNet50 From Scratch & -0.409 & 0.631 & -3.926 & 0.000 & 0.757\\
  ResNet50 Pretrained & -0.940  & 0.213 & -3.601 & 0.000 & 0.776\\
  ResNet101 + LSTM &  0.620 & 0.574 & -3.996 & 0.002 & 0.631 \\
  Bottom-up + LSTM & 0.488  & 0.668 & -4.024 & 0.002 & 0.652\\
  Bottom-up + Transformer & 0.544 & 0.621 & -3.897  & 0.002 & 0.629\\
    \bottomrule
  \end{tabular}
\end{table}

\begin{table}
\tiny
  \caption{Correlation of AUC/SPICE with human judgments of concreteness}
  \label{concreteness-correlation}
  \centering
  \vspace{2mm}
  \begin{tabular}{llllll}
     \toprule

 Model &  Pearson & $p$ & Spearman & $p$\\
 \midrule
  VGG  &   0.303 & 0.195 & 0.277 & 0.238\\
  ResNet50 From Scratch & 0.092 & 0.701 & 0.056 & 0.816\\
  ResNet50 Pretrained &  0.459 & 0.042 & 0.421 & 0.064\\
    ResNet101 + LSTM & 0.733  & 0.001 & 0.598 & 0.011\\
  Bottom-up + LSTM & 0.744 & 0.000 & 0.659 & 0.003 \\
  Bottom-up + Transformer & 0.690 & 0.002 & 0.574  & 0.016 \\
    \bottomrule
  \end{tabular}
\end{table}

\begin{table*}
%\tiny
  \caption{Multiple regression with all variables as predictors of AoA: TalkBank word frequency, AUC/SPICE, and concreteness}
  \label{multiple-regression2}
  \vspace{2mm}
  \centering
  \begin{tabular}{llllllll}
    \toprule
    \multicolumn{8}{c}{Regression Coefficients (predicting AoA)}                   \\
    \cmidrule(r){1-8}
 Model & TalkBank  & $p$ &  AUC / %(classification)  
 & $p$ & Concreteness & $p$ & $R^2$\\ &frequency & &SPICE 
 %(captioning)  
 & & & \\
 \midrule
  VGG Pretrained & -1.111 & 0.096 & -0.015 & 0.986 & -3.747 & 0.000 & 0.794\\
  ResNet50 From Scratch & -1.170 & 0.077 & -0.594 & 0.461  &  -3.704 & 0.000 & 0.801\\
  ResNet50 Pretrained & -0.967 & 0.144 & -0.661 & 0.374 & -3.530 & 0.000 & 0.804\\
  ResNet101 + LSTM &  -1.469 & 0.076 & 0.901 & 0.383 & -4.099 & 0.001 & 0.713\\
  Bottom-up + LSTM & -1.239 & 0.126 & 0.647 & 0.552 & -4.106 & 0.001 & 0.707\\
  Bottom-up + Transformer & -1.445 & 0.082 & 0.782 & 0.447 & -3.951 & 0.001 & 0.709\\ 
    \bottomrule
  \end{tabular}
\end{table*}

Investigation of this result revealed that it is a consequence of substantial collinearity between the computer vision measures and concreteness ratings. The correlations between AUC/SPICE and concreteness are shown in Table \ref{concreteness-correlation}.  The four models that produced statistically significant correlations with AoA are all correlated with concreteness, with statistically significant correlations from all the captioning models.

The observed correlation between concreteness and AUC/SPICE makes sense: concreteness is people's judgment of how well a word corresponds with a visible or tangible thing in the world, and this is what our measures reflect.  A high correlation with concreteness thus indicates that our models are capturing what we intended: the ease of relating a word to its visual referent(s). Importantly, the performance of a computer vision model is an objective quantity that can be estimated directly from a dataset, rather than a subjective quantity that requires additional judgments from people. Further, our approach captures meaningful variability within the visual contexts of concrete and abstract words. For example, the words ``hello'' and ``economy'' may be judged as equally abstract \citep{koper2017improving}, however, ``hello'' may be associated with a more consistent visual context as part of a routine (e.g., waving) compared to ``economy''. Similarly, concrete words like ``spoon'' may have a more consistent surrounding visual context (e.g., a kitchen), compared to words like ``dog'', which may be encountered in many different visual contexts. Future work can apply this approach to go beyond category-level estimates and capture the visual variability of different items, as well as individual differences in the visual contexts that different children experience in densely sampled child-view visual corpora (e.g. \cite{saycam}). By providing a new way to directly measure the concreteness of words, our approach provides a novel metric that can be used in the broader investigation of language processing.

% \begin{figure}
% \begin{center}
% \includegraphics[scale=0.40]{resnet50-fs.png}
% \end{center}
% \caption{\centering\label{fig:resnet50-fs} ResNet50 classification model trained from scratch on COCO: AUC vs child AoA.}
% \end{figure}

% \begin{figure}
% \begin{center}
% \includegraphics[scale=0.40]{resnet50-pretrained.png}
% \end{center}
% \caption{\centering\label{fig:resnet50-pretrained} ResNet50 classification model pretrained on ImageNet: AUC vs child AoA.}
% \end{figure}

% \begin{figure}
% \begin{center}
% \includegraphics[scale=0.33]{resnet101.png}
% \end{center}
% \caption{\centering\label{fig:resnet101} Captioning model using pretrained ResNet101 visual features + LSTM for language: SPICE vs child AoA.}
% \end{figure}

% \begin{figure}
% \begin{center}
% \includegraphics[scale=0.33]{bottomup-lstm.png}
% \end{center}
% \caption{\centering\label{fig:bottomup-lstm} Captioning model using pretrained bottom-up Faster R-CNN visual features + LSTM for language: SPICE vs child AoA.}
% \end{figure}

% \begin{figure}
% \begin{center}
% \includegraphics[scale=0.33]{bottomup-transformer.png}
% \end{center}
% \caption{\centering\label{fig:bottomup-transformer} Captioning model using pretrained bottom-up Faster R-CNN visual features + Transformer for language: SPICE vs child AoA.}
% \end{figure}

%Tsutsui, S., Chandrasekaran, A., Reza, M. A., Crandall, D., & Yu, C. (2020). A Computational Model of Early Word Learning from the Infant's Point of View. Proceedings of the 42nd Annual Conference of the Cognitive Science Society.

\section{Discussion}

We have shown that despite training on only standard machine learning datasets (ImageNet and COCO), several captioning models and one classification model successfully predict the age at which children acquire different categories of words. This result holds across multiple architectures, and for both simple and complex models. This indicates that these models effectively capture the visual difficulty of learning a word for a child. It also suggests that the underlying mechanisms of learning for models and children might be similar in ways that are not yet fully understood but result from the shared statistical structure of the problems they face. 

Figure \ref{fig:regressions} provides some intuition for why visual difficulty goes beyond training data distribution: for example, while the categories ‘food/drink’ and ‘descriptive words’ occur much more frequently in child-directed speech than in the COCO training data, the models are nevertheless successfully predictive of AoA for those categories. This illustrates the value of ML approaches to concreteness, and provides some intuition for the commonalities in child and model learning. Certain categories are also difficult for both models and children, despite those categories being \textit{over}represented in the training data. For example, quantifiers are difficult for both models and children to learn, despite being well represented in COCO training data and child-directed speech. %Cogsci revision: (distributions in the Appendix). 

Pretraining seems to be one of the key differentiating factors between the models which showed substantial correlation and those which did not, such as the ResNet50 model pretrained and the same ResNet50 trained from scratch. %The most noticeable difference here is between the ResNet50 models pretrained and trained from scratch. While the model trained from scratch on our classification task does not show substantial correlation to AoA, the model using pretrained ImageNet weights shows a very substantial correlation. 
There are several potential reasons for this. If pretraining (even on ImageNet alone) supports learning to extract visual features, that skill can be applied to more complex visual features than those in the training data. However, pretraining is not the whole story: the difference between pretrained VGGNet and ResNet50 classification models' correlation to AoA shows that architecture does contribute to the correlation as well. 

The consistently high correlations for captioning models with language components support anecdotal evidence that these larger models combining vision and language modalities do indeed produce more human-like performance for visual word learning. The high correlation across different architectures opens the door to future investigations as to why exactly this is the case -- it is clearly not one particular element, such as a transformer language model, which yields this result. However, sophisticated language modeling with attention mechanisms, present in all the captioning models either through the LSTM or Transformer layers, may be important for producing these results. 

\subsection{Relationship to Previous Work}
%\vfill{-4.3pt}

While there has been no previous work looking directly at predicting AoA from metrics derived from computer vision models, there is an extensive literature in cognitive science and computer vision examining different kinds of correspondences between humans and machines. For example, representations from image classification systems have been used to predict human judgments of image typicality \citep{lake2015deep}, the similarity between images \citep{jozwik2017deep,peterson2018evaluating,hebart2020revealing}, image classification \citep{sanders2020training,battleday2020capturing}, and neural responses to images \citep{yamins2016using,schrimpf2020brain}. Better capturing these aspects of human cognition has been shown to result in improvements in computer vision applications \citep{peterson2019human}. Developmental research has also previously explored the use of deep neural networks to capture aspects of children's language learning, particularly systems that are trained on data from cameras mounted on the heads of infants \citep{bambach2018toddler,orhan2020self,tsutsui2020computational}. This research has primarily focused on visual object learning rather than predicting the timecourse of word-learning itself. Other work has looked at using multimodal neural networks to capture human performance in stylized word-learning settings \citep{vong2022cross}. 
This work provides a converging perspective on how models from computer vision can be used to capture the relationship between linguistic and visual input.

\subsection{Future Work}

In demonstrating how vision and language models' effectively capture word learning difficulty in children, this work also opens the door to more behavioral comparisons of word learning in children and computer vision models. We have demonstrated this result on a standardized group of datasets, with standard pretraining protocols. An important future question is, to what extent particular architectural components (ResNet/Faster R-CNN visual features, or LSTM/Transformer language layers) are important for capturing different facets of child word learning. Is it the scale of larger captioning models which yields the robust similarity to child word learning? Or is it the attention mechanisms in the more sophisticated language components? Another important line of inquiry is how this result changes with datasets; it is surprising that this correlation exists although children and models are certainly exposed to different data. Training models on a child-directed dataset, such as SAYCam \citep{saycam} is likely to strengthen the correlation to child word learning patterns. Our results lay the groundwork for further behavioral comparisons between models and child learning. 

\subsection{Conclusion}
We have shown that the difficulty with which computer vision models learn different categories of words predicts the age at which children learn words in those categories. Although computer vision systems and human children potentially have significant differences in the mechanisms of learning, both face the challenge of relating a word to the visual phenomena it describes. The developmental parallels, which show that the difficulty of learning different categories of words is aligned for both machines and children, suggest that the structure of the learning problem itself may induce similarities in patterns of learning. We hope that these results open the door to new opportunities to model child development using machine learning systems for computer vision and language, and in turn help us to understand these machine learning systems better through their parallels to child development.
%We show this result across different architectural features and different scales of model complexity. We suggest reasons for this similarity, including similar mechanisms of learning under the hood for both models and children. Finally, we suggest further such behavioral comparisons of models and children, and how such work could inform better models of child learning as well as more robust and human-like AI models. 

%TLG: not in anonymized submission
\vspace{1mm}
\begin{small}
\noindent {\bf Acknowledgments.} Mira Nencheva was supported by the ACM SIGHPC Computational \& Data Science Fellowship. This material is based upon work supported by the National Science Foundation (grant number 2107048) and the National Institute of Child Health and Human Development (grant number R01 NICHD 095912). Any opinions, findings, and conclusions or recommendations expressed in this material are those of the authors and do not necessarily reflect the views of these funding agencies. 
\end{small}

\printbibliography

\end{document}